\documentclass{ametsocV6.1}

\usepackage[utf8]{inputenc}
\usepackage[T1]{fontenc} 
\usepackage{siunitx}
\usepackage{multirow}
\usepackage{xcolor}
\usepackage{mathtools}

\newcommand{\var}{\mathrm{Var}}
\newcommand{\E}{\mathbb{E}}

\newcommand{\I}{\mathbb{I}}

\nolinenumbers

\title{Tackling the Accuracy-Interpretability Trade-off in a Hierarchy of Machine Learning Models for the Prediction of Extreme Heatwaves}

\authors{
    Alessandro Lovo,\aff{a,*}
    Amaury Lancelin,\aff{b,c,*}
    Corentin Herbert,\aff{a}
    Freddy Bouchet,\aff{b}\correspondingauthor{Freddy Bouchet, freddy.bouchet@cnrs.fr}
}

\affiliation{
    \aff{a}{ENS de Lyon, CNRS, Laboratoire de Physique, F-69342 Lyon, France}\\
    \aff{b}{LMD/IPSL, CNRS, ENS, Université PSL, École Polytechnique, Institut Polytechnique de Paris, Sorbonne Université, Paris, France}\\
    \aff{c}{RTE France, Paris La Défense, France.} \\
    \aff{*}{Authors contributed equally}
}

\abstract{%
    When performing predictions that use Machine Learning (ML), we are mainly interested in performance and interpretability. This generates a natural trade-off, where complex models generally have higher skills but are harder to explain and thus trust. Interpretability is particularly important in the climate community, especially when dealing with extreme weather events, where gaining a physical understanding of the underlying phenomena is crucial to contain impacts. \\
    In this paper, we perform probabilistic forecasts of extreme heatwaves over France, using a hierarchy of increasingly complex ML models, which allows us to find the best compromise between accuracy and interpretability.
    We use models that range from a global Gaussian Approximation (GA) to deep Convolutional Neural Networks (CNNs), with the intermediate steps of a simple Intrinsically Interpretable Neural Network (IINN) and a model using the Scattering Transform (ScatNet). Our findings reveal that CNNs provide higher accuracy, but their black-box nature severely limits interpretability, even when using state-of-the-art Explainable Artificial Intelligence (XAI) tools. In contrast, ScatNet achieves similar performance to CNNs while providing greater transparency.
    Our interpretable models highlight known drivers of extreme European heatwaves, like persistent anticyclonic anomalies and dry soil, as well as new ones, in the form of sub-synoptic geopotential height oscillations. \\
    This study underscores the potential of interpretability in ML models for climate science, demonstrating that simpler models can rival the performance of their more complex counterparts, all the while being much easier to understand. This gained interpretability is crucial for building trust in model predictions and uncovering new scientific insights.
}

\begin{document}

\maketitle

%
%
%
\statement
    The purpose of this work is to test increasingly complex machine learning models on the task of forecasting extreme heatwaves and explain their prediction. This is important to quantify what additional information the more complex models are able to capture. We find that answering this question with a black-box model and explainability techniques is not sufficient. Indeed, the explanations are mainly qualitative and don't add much to what was known from the simplest linear models. On the other hand, by using an inherently interpretable architecture design, we are able to identify sub-synoptic oscillations in the geopotential height field as the source of additional information, while maintaining the same predictive skill of the black-box model.

%
%

%









\section{Introduction}

Extreme events are responsible for the major part of weather and climate related damages; they have already become one of the most visible hallmarks of climate change over the past decades, and projections indicate that their detrimental impact will keep increasing in the future \citep{IPCC-AR5-extremes}.
Heatwaves, in particular, caused a significant increase in mortality in 2003 in Western Europe \citep{fouilletExcessMortalityRelated2006}, in 2010 in Russia \citep{barriopedroHotSummer20102011} and in 2021 in Canada \citep{hendersonAnalysisCommunityDeaths2022}.
Moreover, the impact of extreme heatwaves extends to losses in the agricultural sector, stress on the power grid, increased wildfire hazards, as well as the endangerment of ecosystems \citep{IPCC-AR6-11}.
It is thus of paramount importance to improve our understanding of these events and to be able to accurately forecast them as early as possible.

Over the past decades, we have witnessed the steady progress of numerical weather prediction over timescales of a few days to a week~\citep{bauerQuietRevolutionNumerical2015}, but also the appearance of skillful probabilistic predictions at subseasonal to seasonal (S2S) timescales, extending beyond the deterministic predictability horizon of the atmosphere.
Typically, S2S prediction systems are based on models of the same type as short range numerical weather prediction, relying on discretized versions of the equations of physics complemented by empirical parameterizations for sub-grid scale phenomena.
While such models do exhibit some skill at prediction of extreme events such as anomalous heat~\citep{Vitart2019, wulff2019higher}, strong limitations remain, in particular for the most extreme events at lead times larger than a few days.
A striking example is that of the 2021 Pacific Northwest heatwave: while state-of-the-art S2S prediction systems succeeded in forecasting above average temperatures several weeks in advance, they severely underestimated the amplitude of the anomaly until about a week before the event~\citep{lin2021WesternNorth2022,domeisenPredictionProjectionHeatwaves2023}.

Another approach to climate prediction is that of statistical models.
Although such models have existed for a long time, they have recently undergone quick developments due to the explosion of computing power and amount of available data which has led to the widespread applications of Machine Learning to the whole field of weather and climate sciences~\citep{braccoMachineLearningPhysics2024}.
In particular, the prediction problem has been addressed with varying degrees of complexity, both in the tools used and in the tasks addressed. For instance, simple neural networks have been used to forecast one dimensional quantities such as heatwave \citep{asadollahPredictionHeatWaves2021,khanPredictionHeatWaves2021} or El Ni\~no \citep{petersikProbabilisticForecastingNino2020} indices. On the other end of the spectrum, foundation models can forecast the global weather with accuracy comparable to state of the art numerical simulations \citep{biAccurateMediumrangeGlobal2023,lamLearningSkillfulMediumrange2023,nguyenClimaXFoundationModel2023}.
Focusing on extreme events, different types of methods have been applied for heatwave prediction over different regions.
For example, \citet{McKinnon2016,Vijverberg2020,Miller2021} have used regression techniques to investigate extreme heat precursors over the US, outperforming dynamical models on timescales up to four weeks.
Similarly, \citet{weirich2023subseasonal} have used linear regression and random forests to forecast heatwaves over Central Europe multiple weeks in advance.
Convolutional neural networks (CNNs) have been applied by~\citet{chattopadhyayAnalogForecastingExtremeCausing2020} and by~\citet{Jacques-Dumas2021,miloshevichProbabilisticForecastsExtreme2023} to the problem of heatwave classification over the US and France, respectively, again exhibiting skill deep into the S2S range.

In spite of these successes, statistical prediction for rare events present a number of challenges \citep{materiaArtificialIntelligenceClimate2024}.
One of the two major shortcomings is the fact that these models require vast amounts of data and extensive resources to be trained, especially the more complex architectures such as deep neural networks which have been increasingly used.
The problem is particularly acute for extreme events, as the events with the highest impact are the rarest, which means there are very few of them in observational records and properly simulating them with climate models is expensive \citep{miloshevichProbabilisticForecastsExtreme2023,Ragone24}.
The second issue is that neural networks essentially behave as black boxes; hence, regardless of their performance, practitioners may struggle to trust them, as they might be making the right prediction for the wrong reason, creating a natural trade-off between performance and interpretability~\citep{rudinStopExplainingBlack2019}.
In addition to consistency with scientific understanding of the underlying phenomena, gaining some degree of interpretability about the predictions would come with other benefits, such as the possibility to improve the models further, to help disentangle correlation and causality, and even to identify unknown mechanisms.

Several strategies have been investigated over the past few years to deal with the lack of data issue.
One is to make better use of the available data, by training the same model on increasingly rare events~\citep{Jacques-Dumas2021} or by fine-tuning a general prediction model to specifically focus on rare events~\citep{Lopez-Gomez2023}.
Another is to use architectures with fewer parameters to learn~\citep{mascoloGaussianFrameworkOptimal2024}.
And finally a third possibility is to generate additional data in a way which samples extreme events preferentially~\citep{Ragone24,ragoneRareEventAlgorithm2021,miloshevichExtremeHeatwaveSampling2023,yiouSimulationExtremeHeat2020}. This latter option opens interesting prospects to improve prediction of rare events, in particular by coupling resampling methods with data-based estimates of the conditional probability of occurrence of these events~\citep{lucenteCouplingRareEvent2022}.

The issue of interpretability has also attracted a growing interest over the past years. There has been an intense proliferation of post-hoc explainability methods \citep{murdochDefinitionsMethodsApplications2019}, mainly imported from the computer science community, which aim at providing the user with human-understandable insight into complex off-the-shelf network architectures.
These methods have had some success in elucidating model predictions for weather and climate~\citep{McGovern2019,tomsPhysicallyInterpretableNeural2020,barnesIdentifyingOpportunitiesSkillful2020}.
However, they are only an approximation of the model’s true prediction \citep{montavonMethodsInterpretingUnderstanding2018,yangInterpretableMachineLearning2024}. Perhaps more importantly, different explanation methods can sometimes yield contrasting results \citep{mamalakisInvestigatingFidelityExplainable2022,mamalakisNeuralNetworkAttribution2022}.
On the contrary, a promising direction is that of developing Machine Learning models which are interpretable by design~\citep{murdochDefinitionsMethodsApplications2019}, and it has been shown that such models can have minimal reduction in performance with respect to complex black box models \citep{barnesThisLooksThat2022,rudinStopExplainingBlack2019,mascoloGaussianFrameworkOptimal2024}.
The drawback is that interpretable models are usually harder to design and train~\citep{murdochDefinitionsMethodsApplications2019,rudinStopExplainingBlack2019,yangInterpretableMachineLearning2024}.\\

In this work, we address the trade-off between prediction accuracy and interpretability in the case of heatwave prediction over France.
Recent work showed that regularized linear regression can be very effective while remaining interpretable~\citep{mascoloGaussianFrameworkOptimal2024,delsoleStatisticalSeasonalPrediction2017,stevensGraphGuidedRegularizedRegression2021}.
Here, we expand on this finding by training a hierarchy of increasingly complex machine learning models, from simple linear regression to fully fledged deep Convolutional Neural Networks.
We explore two ways to build intermediate rungs in the hierarchy and the consequences in terms of the accuracy-interpretability trade-off: linear feature extraction followed by nonlinear prediction, or nonlinear feature extraction followed by linear regression (section~\ref{sec:models}).
We compare the performance of the four models, both with a dataset characteristic of the largest ones available from climate models, and with a much shorter one, more typical of observations (section~\ref{sec:performance}).
Finally, we compare built-in interpretability of the two methods based on linear feature extraction with post-hoc explainability of the more complex networks (section~\ref{sec:interpretability}).
A first aim is purely methodological.
Additionally, we try to understand what \emph{extra} information these latter models are able to capture with respect to their simpler counterparts.
We show that linear regression already performs well and identifies commonly accepted heatwave drivers like persistent anticyclones and dry soil.
CNNs have a higher performance, but post-hoc explanations provide little physical insight beyond linear regression.
On the contrary, an interpretable variant of Scattering networks \citep{brunaInvariantScatteringConvolution2013} achieves the same skill as the CNNs, but  pinpoints much more precisely the source of improvement with respect to the linear regression baseline, finding extra predictability in sub-synoptic oscillations in the geopotential height field.


\section{Data and methods}

\subsection{Data}
    We use the 1000 year-long control run of the Community Earth System Model (CESM) version 1.2.2 \citep{hurrellCommunityEarthSystem2013}, already used in \citet{ragoneRareEventAlgorithm2021}. Only atmosphere and land components are dynamic, while sea surface temperature, sea ice and greenhouse gas concentrations are prescribed to reproduce a stationary climate that resembles the one of the year 2000.
    The model is run with a $\SI{0.9}{\degree}$ resolution in latitude, $\SI{1.25}{\degree}$ in longitude and 26 pressure levels. Previous studies \citep{ragoneRareEventAlgorithm2021,miloshevichRobustIntramodelTeleconnection2023} have shown that, with this setup, CESM is able to accurately reproduce the relevant atmospheric phenomena connected to heatwaves, for instance planetary teleconnection patterns \citep{miloshevichRobustIntramodelTeleconnection2023}.

    Since we focus on summer heatwaves, we will use daily averaged data for the months of June, July, and August.

    Of the total 1000 years of data available, we keep the last 200 for testing, while the first 800 are split in training and validation according to a 5-fold cross validation process.

\subsection{Heatwave amplitude}
    In the literature, there are many definitions of what a heatwave is \citep{HeatwaveDefinitions}, with most of them involving hard thresholds that need to be overcome for a specific amount of consecutive days.
    In this paper, we follow the definition used in \citet{Ragone24,galfiLargeDeviationTheorybased2019,galfiFingerprintingHeatwavesCold2021,ragoneRareEventAlgorithm2021,jacques-dumasDatadrivenMethodsEstimate2023,miloshevichProbabilisticForecastsExtreme2023,mascoloGaussianFrameworkOptimal2024}, which gives a continuously varying heatwave amplitude $A$ and allows easily to control the heatwave duration and intensity.

    We define the heatwave amplitude at $A(t)$ as
    \begin{equation} \label{eq:heatwave}
        A(t) := \frac{1}{T}\int_{t}^{t+T} \left( \frac{1}{\mathcal{A}}\int_{\mathcal{A}}T_\text{2m}(\vec{r},u)d\vec{r}  \right)du,
    \end{equation}
    where $T$ is the heatwave duration in days, $\mathcal{A}$ is the geographical region of interest, and $T_\mathrm{2m}$ is the \SI{2}{\meter} temperature anomaly, defined by subtracting the multi-year mean of temperature of each day.

    In this work, $\mathcal{A}$ will be the region of France, which has a size of roughly \SI{1000}{\kilo\meter} and thus sits nicely at the scale of cyclones and anticyclones, which is the relevant one for large-scale atmospheric dynamics. Moreover, studying heatwaves involving a whole country can be very relevant for policymakers \citep{barriopedroHotSummer20102011,fouilletExcessMortalityRelated2006}.
    We focus here on two-week heatwaves ($T=14$), as longer lasting heatwaves have a much higher impact than shorter ones \citep{IPCC-AR5-extremes,IPCC-AR6-11,meehlMoreIntenseMore2004,amengualProjectionsHeatWaves2014}.

    Note that since $A(t)$ contains a 14-day running mean, the heatwave amplitude of consecutive days will be highly correlated.

\subsection{Predictors}
    To forecast a heatwave that starts at time $t$, we will use a set of predictors $X(t-\tau)$, where $\tau$ is the lead time. Since our data has only land and atmosphere as dynamical components, we will neglect ocean variables like sea surface temperature or sea ice, although in the real world they can be important sources of predictability for heatwaves at the subseasonal to seasonal timescale \citep{HeatwaveDefinitions,domeisenPredictionProjectionHeatwaves2023}.
    We will thus use as predictors the geopotential height anomaly at \SI{500}{\hecto\pascal} (Z500) over the whole Northern Hemisphere and soil moisture anomalies (SM) over France, which have been shown to be the most informative weather fields \citep{miloshevichProbabilisticForecastsExtreme2023}.

    Indeed, while the geopotential height anomaly at the middle of the troposphere gives good insight into the wind flow (thanks to the geostrophic approximation), soil moisture acts as an important modulator of the likelihood of heatwaves at mid-latitudes, as it controls the evaporative cooling potential of the surface \citep{HeatwaveDefinitions,miloshevichProbabilisticForecastsExtreme2023,bensonCharacterizingRelationshipTemperature2021,dandreaHotCoolSummers2006,fischerSoilMoistureAtmosphere2007,hirschiObservationalEvidenceSoilmoisture2011,lorenzPersistenceHeatWaves2010,rowntreeSimulationAtmosphericResponse1983,schubertNorthernEurasianHeat2014,shuklaInfluenceLandSurfaceEvapotranspiration1982,stefanonHeatwaveClassificationEurope2012,vargaszeppetelloProjectedIncreasesMonthly2020,zeppetelloPhysicsHeatWaves2022,zhouLandAtmosphereFeedbacks2019,vautardSummertimeEuropeanHeat2007}.
    To facilitate the work of neural networks, each grid point of each of the two weather fields is independently standardized to zero mean and unitary standard deviation
    (see section~S.5 of the Supplementary Materials).

    Since this study is mainly methodological, we will focus on a single value of the lead time $\tau$. For simplicity, we chose $\tau=0$, which means we'll be predicting the average temperature of the coming two weeks.

\subsection{Probabilistic regression}
    We aim at predicting the distribution $\hat{p}(A|X)$ of the heatwave amplitude $A$, given the current state of the predictors $X$.
    From this probabilistic regression, we can build a classifier which computes the conditional probability of the heatwave amplitude exceeding a given threshold $a$: $q(X) = \mathbb{P} (A > a|X)$.
    This quantity, called the committor function~\citep{lucenteCommittorFunctionsClimate2022}, was used previously as the target for heat wave prediction with machine learning methods~\citep{miloshevichProbabilisticForecastsExtreme2023,mascoloGaussianFrameworkOptimal2024}.

    Here, we use three metrics to quantify the quality of our prediction: two for a regression task, and one for the task of classifying the 5\% most extreme events.
    The first two are the Negative Log Likelihood (NLL)
    \begin{equation}
        NLL = -\frac{1}{N} \sum_{i=1}^N \log \hat{p}(A_i|X_i),
    \end{equation}
    and the Continuous Ranked Probability Score (CRPS)
    \begin{equation}
        CRPS = \frac{1}{N} \sum_{i=1}^N \int_{-\infty}^{+\infty} \left(\I_{a \geq A_i} - \int_{-\infty}^{a} \hat{p}(a^\prime|X_i) da^\prime \right)^2 da,
    \end{equation}
    where $\I$ is the indicator function~\citep{Wilks2019}.

    The third one is the Binary Cross Entropy (BCE)
    \begin{equation}
        BCE = -\frac{1}{N} \sum_{i=1}^N \left( \I_{A_i < a_5}\log \left( \int_{-\infty}^{a_5} \hat{p}(a^\prime|X_i) da^\prime\right) + \I_{A_i \geq a_5}\log \left( \int_{a_5}^{+\infty} \hat{p}(a^\prime|X_i) da^\prime\right) \right),
    \end{equation}
    where $a_5 = \SI{3.11}{\kelvin}$ is the threshold that defines the 5\% most extreme heatwaves over the training set. The index $i$ iterates through all daily snapshots of data either in the 200-year-long test set, the five 640-year-long training sets or the five 160-year-long validation sets.

    To have a reference, we use the zero-order prediction given by the climatology $\hat{p}_\mathrm{clim}(A)$ which does not use any information about the predictors, but rather fits the distribution of $A$ on the training set and uses it to compare with the data in the validation and test sets. For the first two metrics, we compute $\hat{p}_\mathrm{clim}(A)$ using a kernel density estimate, while for the BCE, we simply use the fact that $\int_{a_5}^{+\infty} \hat{p}_\mathrm{clim}(a^\prime) da^\prime = 0.05$.

    For a prediction performed by a $\{MODEL\}$, we compute the skill score of a $\{METRIC\}$ as the relative gain from climatology:
    \begin{equation}\label{eq:skill}
        \{METRIC\}S = 1 - \frac{\{METRIC\}_{\{MODEL\}}}{\{METRIC\}_\mathrm{clim}}
    \end{equation}
    A perfect prediction has a skill score of $1$, while a prediction that is worse than the climatology results in negative values. The former would mean that the predicted probabilities $\hat{p}(A|X_i)$ are Dirac's deltas on the values $A = A_i$ that actually happen in the data, so we don't expect to reach skills that are close to 1.


\section{The model hierarchy}\label{sec:models}

To perform probabilistic regression, we will approximate the conditional distribution of the heatwave amplitude $\hat{p}(A|X)$ as a Gaussian distribution with mean $\hat{\mu}(X;\theta)$ and variance $\hat{\sigma}^2(X;\theta)$, as was done for instance for El Niño~\citep{petersikProbabilisticForecastingNino2020}. We will use increasingly complex models to parameterize $\hat{\mu}$ and $\hat{\sigma}$. A summary of this model hierarchy is presented in Tab.~\ref{tab:hierarchy}.

\begin{table}[htbp]
    \centering
    \begin{tabular}{c|cc|ccc}
        \multirow{2}{*}{Method} & \multirow{2}{*}{$\hat{\mu}(X;\theta)$} & \multirow{2}{*}{$\hat{\sigma}(X;\theta)$} & trainable & non-trainable & \multirow{2}{*}{hyperparameters} \\
        & & & parameters & parameters & \\
        \hline
        GA & $M\cdot X$ & $\sigma$ & 27 425 & 0 & 1 \\
        IINN & $g_\mu(M\cdot X)$ & $s(g_\sigma(M\cdot X))$ & 55 058* & 0 & 10 \\
        ScatNet & $\beta_\mu\cdot \phi(X)$ & $s(\beta_\sigma \cdot \phi(X))$ & 19 930* & 656 640* & 5 \\
        CNN & $g_\mu(X)$ & $s(g_\sigma(X))$ & 684 000* & 0 & 10
    \end{tabular}
    \caption{Complexity of the hierarchy of probabilistic prediction models. Values marked with * denote the number of parameters at the optimum with respect to hyperparameters. $M$ is the projection pattern, $g_\mu$ and $g_\sigma$ are general non-linear functions parameterized by neural networks, $\phi$ is the scattering transform, $\beta_\mu$ and $\beta_\sigma$ are projection patterns in the transformed space, and $s$ is the softplus function ($s(x) = \log(1 + \exp(x))$), which ensures that $\hat{\sigma}(X;\theta) > 0$.}
    \label{tab:hierarchy}
\end{table}

\subsection{Gaussian approximation}
    The simplest option after the climatology is to perform a linear regression of $A$ against $X$. The underlying assumption of this method is that the joint distribution of $X$ and $A$ is a multivariate Gaussian \citep{mascoloGaussianFrameworkOptimal2024}, and results in a prediction with linear mean $\hat{\mu}(X;\theta) = M\cdot X$ and constant standard deviation $\hat{\sigma}(X;\theta) = \sigma$.
    $M$ has the same dimensions as $X$ and is an optimal projection pattern, that condenses all the important information for heatwave prediction into the scalar index $F = M\cdot X$. Training is performed in one step, with
    \begin{equation}\label{eq:ga}
        \begin{cases}
            M &= \arg\min_{M} \left(\frac{1}{N} \sum_{i=1}^N \left( A_i - M\cdot X_i \right)^2 + \epsilon H_2(M) \right) \\
            \sigma^2 &= \var[A] - \frac{(\E[FA])^2}{\var[F]}
        \end{cases} ,
    \end{equation}
    where $\E[\bullet]$ and $\var[\bullet]$ denote respectively expectation and variance of $\bullet$, computed on the training set.
    For high dimensional climate data, we need to regularize the projection pattern $M$, and this is achieved by penalizing the norm of its spatial gradient $H_2(M)$, which forces $M$ to be spatially smooth. $\sigma^2$ is the residual variance of $A$ once we remove the linear trend $M\cdot X$. See \citet{mascoloGaussianFrameworkOptimal2024} for more details.

    For this method the trainable parameters are $\theta = \{M,\sigma\}$ and the single hyperparameter is the regularization coefficient $\epsilon$.

\subsection{Intrinsically Interpretable Neural Network}
    To go a step further from the Gaussian approximation, we still project onto an optimal index $F = M \cdot X$, but then we feed it to a (relatively small) fully connected neural network to compute $\hat{\mu}$ and $\hat{\sigma}$. We call this method Intrinsically Interpretable Neural Network (IINN) \citep{lovoInterpretableProbabilisticForecast2023}, as the prediction is decomposed in a linear projection $M\cdot X$, of which we can visualize the projection pattern $M$ (as for GA), and two non-linear scalar functions $g_\mu(M\cdot X)$ and $s(g_\sigma(M\cdot X))$, which are as well very easy to visualize.

    In this case, training is performed by gradient descent of the CRPS loss with the added regularization $+\, \epsilon H_2(M)$. The trainable parameters are the components of $M$ and the weights of the network, while hyperparameters are $\epsilon$, the architecture of the network (number of layers and neurons per layer), plus the usual hyperparameters concerning training, like learning rate and batch size.

\subsection{Scattering Network}
    The \emph{Scattering Transform} was introduced by \citet{mallat2012group} as an alternative to traditional convolutional neural networks (CNNs).
    They aim to create a representation of the data that is stable to local deformations and translations while still preserving essential information. It computes a set of features by applying convolutions with a set of \emph{fixed} wavelet filters to the input signal (such as an image).  Given this set of features, we can then apply a classifier or a regressor to obtain the final prediction, depending on the task we are interested in. The regressor could be a simple linear layer, since the features are already extracted. One can also combine the Scattering Transform with an extra CNN on top of it, as it is done in \citet{oyallonScatteringNetworksHybrid2019}.\\

    Since its introduction, the Scattering Transform has shown promising results for various physical fields and tasks \citep{cheng2024scattering}, especially when the data have limited training samples or when the task requires robustness to deformations and translations. It has been applied, for instance, to quantum chemical energy regression and the prediction of molecular properties \citep{eickenberg2018solid}, and to astrophysics through the statistical description of the interstellar medium \citep{allys_rwst_2019}. This work is, to the best of our knowledge, the first time that scattering networks are applied to atmospheric dynamics. \\

    The main steps of the scattering transform are as follows:
    \begin{itemize}
        \item Wavelet Transform and non-linearity: the input signal is convolved with a set of wavelet filters to extract frequency and phase information at different scales. The set of wavelet filters is arranged to tile the Fourier space, which is discretized by scales and orientations. Consequently, each wavelet filter is located in a distinct position in Fourier space, defined by a scale $j \in \{1,\dots,J\}$ and an orientation $l \in \{1,\dots,L\}$. The result of the wavelet transform is then passed through a point-wise non-linear modulus operator.

        \item Pooling and Aggregation: To achieve local translation invariance, a downsampling operation is applied to the transformed coefficients. This helps in reducing the spatial resolution of the representation while retaining the essential information.  We apply a local averaging operator (typically a Gaussian smoothing function), followed by an appropriate downsampling by a factor $2^J$.
        \item Recursive Operation: The same wavelet transform, non-linearity, and pooling steps can be repeated a second time to create a two-layered scattering representation. Each layer captures different levels of abstraction and invariance to transformations. It is not useful to compute higher order scatterings, because their energy is negligible \citep{brunaInvariantScatteringConvolution2013}.
    \end{itemize}
    We used the implementation offered by the python package \emph{Kymatio} \citep{andreux2020kymatio}. \\

    Here we predict $\hat{\mu}(X;\theta)$ and $\hat{\sigma}(X;\theta)$ by simply applying a fully connected layer to the (flattened) features obtained from the concatenation of the Scattering Transform of the \SI{500}{\hecto\pascal} geopotential height anomaly field and the raw pixels of soil moisture over France. We call the resulting model \emph{ScatNet}.
    Note that when we refer to the \emph{scattering transform}, we are specifically referring to the first (unsupervised) part of the model, excluding the fully connected layer.
    The architecture hyperparameters to tune are the number of scales $J$ and orientations $L$ at each scale for the wavelet filters, and the maximum order of the scattering (the number of recursive applications of the wavelet transform, non-linearity, and pooling steps), which can be either 0, 1, or 2  \citep[see][]{brunaInvariantScatteringConvolution2013}. After hyperparameter optimization, we fixed $J=3$, $L=8$, and the maximum order to 1. The hyperparameters related to the training phase are the learning rate and the batch size.

\subsection{Convolutional Neural Network}
    The most complex model in this study is a Convolutional Neural Network (CNN). In this case, we follow a similar setup to \citet{miloshevichProbabilisticForecastsExtreme2023}, where we feed the input $X$ to the neural network as a two-channel image, with the two `colors' corresponding to the geopotential height and soil moisture fields, the latter being set to 0 outside France. This `image' is processed by several convolutional layers, then by several fully connected ones, with the final output being $\hat{\mu}$ and $\hat{\sigma}$.
    The network is trained by gradient descent of the CRPS loss, and the hyperparameters of this model are its architecture plus learning rate and batch size.

\subsection{Hyperparameter optimization}
    Hyperparameters of IINN, ScatNet and CNN are optimized using a Bayesian search algorithm provided by the \emph{optuna} \citep{optuna} python package. The best combination is the one that gives the highest validation Binary Cross Entropy Skill (BCES) (see Eq.~\eqref{eq:skill}). We also tried optimizing with respect to the other metrics and the results do not change significantly.

    The regularization coefficient $\epsilon$ of GA and IINN is treated separately, as it controls how smooth, and thus interpretable, the projection pattern $M$ is.
    At the best configuration of hyperparameters found by \emph{optuna}, we systematically try logarithmically spaced values of $\epsilon$. The skills of both GA and IINN display a broad plateau for intermediate values of $\epsilon$, with poor performance for either too high or too low values
    (see Figs.~S1~and~S2 of the Supplementary Materials).
    We then choose $\epsilon$ as the one that leads to the smoothest projection pattern without loss of performance.


\section{Performance}\label{sec:performance}

\subsection{Training on the full dataset}

    In Tab.~\ref{tab:skill}, we show the performance of the hierarchy of models on the test set at the optimal values for hyperparameters. Since at the end of the k-fold cross validation process we have 5 trained models, we evaluate all of them on the test set, which allows us to obtain error bars on the skill of the networks. Note that since the training sets of the 5 folds share a significant portion of the data, while the validation sets are independent, the error mainly quantifies variance due to different initialization of the networks and validation on different datasets.

    As expected, the more complex architectures perform better than the simple ones, but GA and IINN are very close, showing that once we project the high dimensional predictors $X$ onto the scalar variable $F = M\cdot X$, allowing for non-linearity in the forms of $\hat{\mu}(F)$ and $\hat{\sigma}(F)$ does not improve significantly the performance.
    Similarly, ScatNet and CNN have very comparable skills, suggesting that the important part of the prediction is capturing the local structures in the data, which can be achieved as easily by a deterministic wavelet transform as with more complex learned convolutional filters. In this case too, after the important features are extracted, there appears to be no benefit in the more complex parametrization of $\hat{\mu}$ and $\hat{\sigma}$ provided by the CNN.

    \begin{table}[htbp]
        \centering
        \begin{tabular}{c|c|c|c|c|}
			\multicolumn{2}{c}{} & \multicolumn{3}{c}{Metric} \\
			\cline{3-5}
			\multicolumn{2}{c|}{} & CRPSS & NLLS & BCES \\
			\cline{2-5}
			\multirow{4}{*}{\rotatebox[origin=c]{90}{Model}}
			 & GA &$0.2864 \pm 0.0009$ & $0.2169 \pm 0.0009$ & $0.293 \pm 0.001$ \\
			\cline{2-5}
			 & IINN &$0.287 \pm 0.002$ & $0.217 \pm 0.002$ & $0.291 \pm 0.003$ \\
			\cline{2-5}
			 & ScatNet &$0.3097 \pm 0.0007$ & $\mathbf{0.246 \pm 0.003}$ & $\mathbf{0.314 \pm 0.005}$ \\
			\cline{2-5}
			 & CNN &$\mathbf{0.310 \pm 0.003}$ & $0.245 \pm 0.007$ & $0.311 \pm 0.008$ \\
			\cline{2-5}
		\end{tabular}
        \caption{Test skills (the higher, the better) of the different models, shown as mean and standard deviation over the 5 folds. In bold, the best performing model according to each of the three metrics.}
        \label{tab:skill}
    \end{table}

\subsection{Training on a smaller dataset}
    The previous section showed that the most accurate predictions are obtained with the more complex models when training on 640 years (and validating on 160).
    Often, climate datasets are much shorter, and machine learning techniques are notoriously data-hungry.
    Thus, we perform a second experiment using a total of 80 years (size comparable to the reanalysis datasets), 64 for training and 16 for validation, as usual optimizing hyperparameters to maximize validation skill.

    When we test the trained models on the same 200-year-long test set as before, we obtain the results presented in Tab.~\ref{tab:skill-80y}, which show a reversal of the ranks, with the Gaussian approximation clearly outmatching all other methods. In \citet{mascoloGaussianFrameworkOptimal2024} the authors already point out the remarkable robustness of the Gaussian approximation to lack of data, but here we see that even the relatively similar method of the IINN suffers a lot from the dataset size, especially for the classification task, with a BCES comparable to the CNN's.

    The main reason for this seems to be the optimization method, as GA is trained in one step (see Eq.~\eqref{eq:ga}), while all other methods use stochastic gradient descent, which in a lack of data regime may be prone to overfitting, both during training and during the hyperparameter optimization phase
    (see Fig.~S2 of the Supplementary Materials).
    To test this hypothesis, we created a variant of ScatNet where $\hat{\mu}(x)$ is estimated with a direct linear regression from the scattering coefficients (minimizing mean square error in one step) and $\hat{\sigma}$ is kept constant at the value estimated by GA. And indeed, with this setup we were able to achieve skills that are a few percentage points better than GA (not shown here).




    \begin{table}[htbp]
        \centering
        \begin{tabular}{c|c|c|c|c|}
            \multicolumn{2}{c}{} & \multicolumn{3}{c}{Metric} \\
            \cline{3-5}
            \multicolumn{2}{c|}{} & CRPSS & NLLS & BCES \\
            \cline{2-5}
            \multirow{4}{*}{\rotatebox[origin=c]{90}{Model}}
                & GA &$\mathbf{0.250 \pm 0.004}$ & $\mathbf{0.165 \pm 0.006}$ & $\mathbf{0.262 \pm 0.006}$ \\
            \cline{2-5}
                & IINN &$0.241 \pm 0.008$ & $0.14 \pm 0.02$ & $0.22 \pm 0.02$ \\
            \cline{2-5}
                & ScatNet &$0.230 \pm 0.004$ & $0.05 \pm 0.01$ & $0.227 \pm 0.002$ \\
            \cline{2-5}
                & CNN &$0.22 \pm 0.02$ & $0.09 \pm 0.03$ & $0.22 \pm 0.03$ \\
            \cline{2-5}
        \end{tabular}
        \caption{Test skills (the higher, the better) of the different models when trained on the smaller 80 year dataset, shown as mean and standard deviation over the 5 folds. In bold, the best performing model according to each of the three metrics.}
        \label{tab:skill-80y}
    \end{table}

\subsection{A remark on regression and classification}
    In this work, the different neural network architectures were trained to minimize the CRPS loss, but virtually identical skills can be obtained minimizing the NLL loss. Similarly, the Gaussian approximation minimizes a regression metric (see Eq.~\eqref{eq:ga}).
    On the other hand, training on BCE yields a worse performance (even for the BCES metric itself). Indeed, training on a classification task separates the training data in two classes (heatwave and non-heatwave), neglecting the information about the heatwave amplitude. For the particular problem of heatwaves over France, the choice of the threshold $a_5$ to distinguish the two classes is arbitrary, and there is no obvious regime shift between mild and very extreme heatwaves \citep{noyelleInvestigatingTypicalityDynamics2024,mascoloGaussianFrameworkOptimal2024}.


\section{Interpretability}\label{sec:interpretability}

Now that we know how the different models in the hierarchy perform, we will try to build an understanding of how they provided their predictions. The goal is twofold: on the one hand, relating properties of the input to the specific components of the statistical models may help improving the performance of the method, and on the other hand it may contribute to understanding the relative importance of physical processes for predictability.

\subsection{GA and IINN}\label{sec:ga-projmap}
\begin{figure}[htbp]
        \centering
        \includegraphics[width=1.\textwidth]{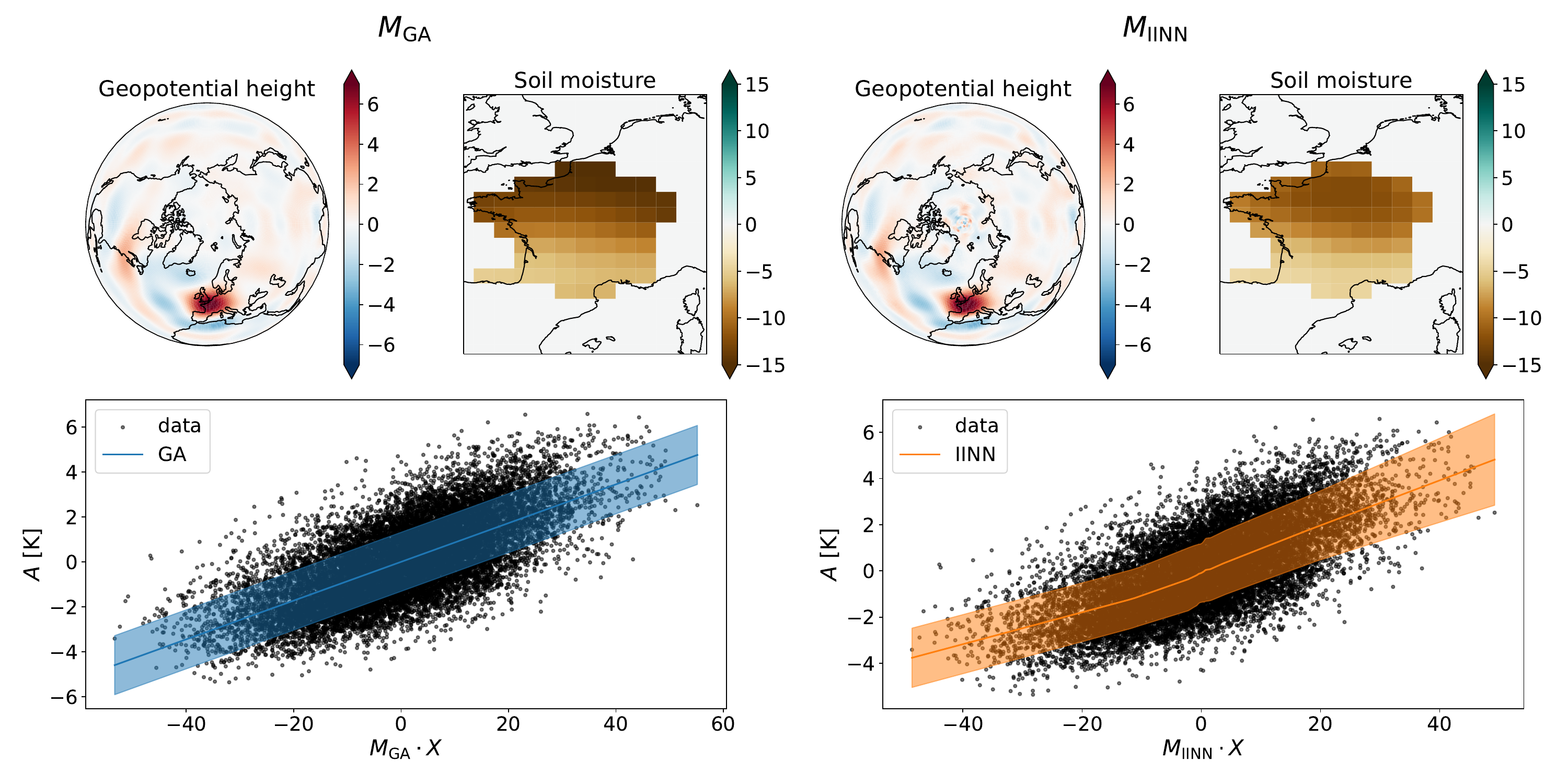}
        \caption{Projection patterns (top) and projected space (bottom) for GA (left) and IINN (right). In the bottom plots, the black dots are the test data, the continuous line is the predicted $\hat{\mu}(X)$ and the shading corresponds to $\pm \hat{\sigma}(X)$. We show among the 5 models the one with the highest skill.}
        \label{fig:ga-iinn}
    \end{figure}

    Both GA and IINN are fully interpretable from a methodological standpoint: they perform a linear projection with a pattern $M$ followed by scalar functions of the index $F = M\cdot X$.
    In Fig.~\ref{fig:ga-iinn}, we plot the projection patterns $M$ and the predictions $\hat{\mu}(F), \hat{\sigma}(F)$ in the projected space.

    For both methods, the functions $\hat{\mu}(F)$ increase with the index $F$ (Fig.~\ref{fig:ga-iinn}, bottom panels), meaning that geopotential and soil moisture anomaly maps positively correlated with the projection patterns $M$ increase the probability of extreme temperatures over France.
    We note that, by construction, the Gaussian approximation performs a linear fit of the data with constant predicted variance.
    While the IINN does not have any a priori constraint on the shape of these functions, we observe that $\hat{\mu}(F=M_\mathrm{IINN}\cdot X)$ is close to linear, with only slightly different slopes depending on the sign of $F$, and exhibits a slightly larger variance for large values of the index.
    However, the values of the two slopes are not very robust across the 5 folds, sometimes being very similar, and as well the trend in the predicted variance is not stable.

    We now consider the projection patterns $M$ to discuss the physical interpretation of the predictions (Fig.~\ref{fig:ga-iinn}, top panels).
    These patterns are very robust across the folds and look very similar for GA and IINN, the main difference being a greater relative importance of soil moisture in the former.
    The most prominent features of the projection patterns are the negative weights for soil moisture, and the large positive coefficients for geopotential over Western Europe.
    This is in agreement with the understanding that persistent anticyclonic anomalies and dry soil are important predictors of high temperatures in the mid-latitudes \citep{HeatwaveDefinitions,ardilouze2017multi,weirich2023subseasonal,barriopedroHeatWavesPhysical2023,miloshevichProbabilisticForecastsExtreme2023}.
    Note that both models attribute larger weight to soil moisture in Northern France than in Southern France, highlighting greater predictability from the former.
    Another notable property of the projection map for geopotential is the presence of a coherent pattern over the North Atlantic, with positive anomalies over the storm tracks region and negative anomalies over the ocean contributing to the prediction of high temperatures, suggesting anomalies in the path or strength of the Jet Stream.
    However, the projection map does not allow us to distinguish between different mechanisms pointed out previously, such as stationary Rossby waves \citep{screen2014amplified,petoukhov2013quasiresonant,dicapuaDriversSummer20102021,ragoneRareEventAlgorithm2021,miloshevichRobustIntramodelTeleconnection2023} or transient eddies \citep{lehmann2015influence}.

    In conclusion, the interpretation of the projection patterns is consistent with the physical understanding of heatwave mechanisms, and thus gives us trust in the statistical models.
    However, the patterns do not reveal new mechanisms, and do not allow us to discuss the relative role of the specific dynamical mechanisms associated with geopotential anomaly patterns.
    Finally, we note that the slightly larger complexity of the IINN does not alter the projection patterns, and the two methods behave in a very similar way.

    \subsection{CNN}
        The previous methods were simple enough that it is possible to relate the trained weights to the final prediction in a rather intuitive way, which allowed us to explain the prediction at once for any input $X$ (global interpretability). This is not the case for the CNN, which behaves like a black box and thus is not methodologically interpretable. However, we can still gain physical interpretability by using post-hoc explainable AI techniques, and most will give a different explanation for each input $X_i$ (local interpretability).
        These methods are slightly sensitive to the weights of the neural network, and so when we applied them to the 5 CNN models, we obtained subtly different results. However, the qualitative conclusions are the same, so, for simplicity, we show the results only for one of the 5 folds.

        \subsubsection{Global interpretability: Optimal Input}
\begin{figure}
                \centering
                \includegraphics[width=0.8\textwidth]{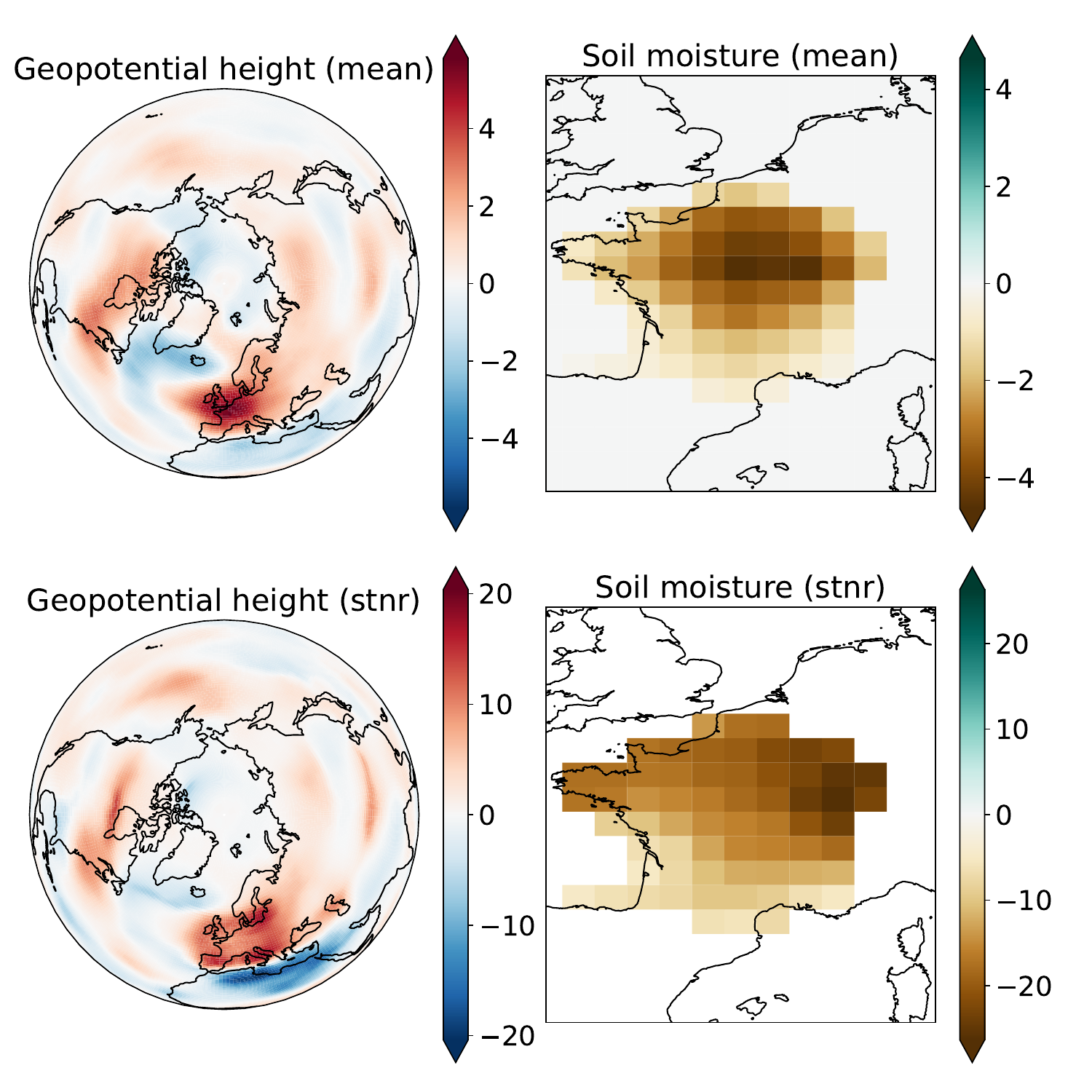}
                \caption{Top: average of the optimal inputs $S$ that maximize the heatwave amplitude $\hat{\mu}_\mathrm{CNN}(S)$ predicted by the CNN, across the different seeds $S_0$ taken from the test dataset. Bottom, signal-to-noise-ratio, computed as the pixelwise ratio between mean (top row) and standard deviation of the optimal inputs (see Fig.~S11 of the Supplementary Materials).}
                \label{fig:oi}
            \end{figure}

            A possibility to investigate the prediction of the CNN is computing the input $S$ (i.e. a map of \SI{500}{\hecto\pascal} geopotential height and soil moisture anomalies) that, once fed to the CNN, yields the highest predicted heatwave amplitude.
            This method, referred to as Backward Optimization \citep{olahFeatureVisualization2017}, has been showcased for various problems in weather and climate, such as tornado and hail prediction \citep{McGovern2019} or identification of ENSO phases and seasonal prediction of continental temperatures in North America \citep{tomsPhysicallyInterpretableNeural2020}.
            A naïve application of the method consists in initializing our input $S_0$ with a given snapshot from the test data and then performing gradient descent of the loss $\ell = -\hat{\mu}_\mathrm{CNN}(S)$, not on the weights of the CNN but on the coordinates of $S$.
            However, doing so yields very noisy maps and absurdly high predicted heatwave amplitudes
            (see Fig.~S7 of the Supplementary Materials).
            Hence, we add regularization terms to constrain the L2 norm $|S|$ and roughness $\sqrt{H_2(S)}$ of the optimized input $S$:
            \begin{equation}
                \ell = -\hat{\mu}_\mathrm{CNN}(S) + \lambda_2 \left(|S| - n_0 \right)^2 + \lambda_r \left(\sqrt{H_2(S)} - r_0\right)^2 ,
            \end{equation}
            where $n_0=0.7$ and $r_0=28$ were empirically chosen as the average L2 norm and roughness of the test data
            (see Fig.~S8 of the Supplementary Materials).
            We searched logarithmically spaced values of the two regularization coefficients, settling on $\lambda_2 = 100$ and $\lambda_r = 0.1$, which were the smallest values ensuring that L2 norm and roughness of the optimized input $S$ were close to $n_0$ and $r_0$, respectively.
            We then run the optimization initializing with all the snapshots in the test dataset, which gives us slightly different optimized inputs due to the non-convex nature of the optimization.
            In the top row of Fig.~\ref{fig:oi}, we show the average optimized input, and we note that in spite of the constraints in the loss, the Z500 optimal input map does not resemble \emph{physical} geopotential anomalies, as most of the mass is concentrated over the Atlantic sector.
            Indeed, the optimization procedure yields heatwave amplitudes which are much larger than any value in the dataset, with an average $\hat{\mu}_\mathrm{CNN}(S) = 11.5 \pm 0.2 \si{\kelvin}$ (see
            Fig.~S9 of the Supplementary Materials
            for the full histogram).

            Nevertheless, the average optimal input looks very similar to the projection pattern of the Gaussian approximation (top left panel of Fig.~\ref{fig:ga-iinn}), with the notable difference that it attributes a smaller weight to the positive geopotential anomaly over Western Europe and that this anomaly extends further North over Scandinavia.
            The latter property is reminiscent of the findings of \citet{della-martaSummerHeatWaves2007} who report statistical connections between high pressure in this area and heatwaves over Western Europe in historical data.
            It also hints at a connection to the positive NAO and Atlantic low regimes identified by \citet{cassou2005tropical}.

            In the bottom panel of Fig.~\ref{fig:oi}, we plot the signal-to-noise-ratio (STNR) defined as the pixelwise ratio between mean and standard deviation of the optimal inputs (see
            Fig.~S11 of the Supplementary Materials).
            The very high values of STNR indicate the robustness of the average optimal input pattern. In particular, there is a very strong convergence (i.e. high STNR) of the optimized inputs over Europe and Northern Africa, hinting at the importance that the main anticyclone doesn't extend too far South but rather stays concentrated over Central Europe.
            In the teleconnection patterns over the Atlantic, North America and Siberia, there is also a quite strong convergence around thin zonal lines, but this may be an artifact due to the architecture of the CNN.
            On the other hand, the Arctic region displays particularly small values of the STNR, due to an amplification of the original unitary variance initially present in the seeds $S_0$ (see
            Fig.~S11 of the Supplementary Materials).
            This may suggest that in this region the information in the geopotential height field is used non-linearly by the CNN.

            Concerning soil moisture, the STNR is greatest in the Northeastern corner of France, despite the maximum in the mean being rather in the center, but this is not a robust pattern across the different folds.


            To explain why the optimal input resembles the projection pattern of the Gaussian approximation, we can write the prediction of the CNN as a perturbation on top of the prediction of the GA:
            \begin{equation}
                \hat{\mu}_\mathrm{CNN}(X) = \hat{\mu}_\mathrm{GA}(X) + \left(\hat{\mu}_\mathrm{CNN}(X) - \hat{\mu}_\mathrm{GA}(X)\right) = M_\mathrm{GA} \cdot X + \hat{\mu}_\mathrm{pert}(X) .
            \end{equation}
            Since the GA gives already very good predictions, $\hat{\mu}_\mathrm{pert}(X)$ will be a small non-linear perturbation, and indeed, as can be seen by
            Fig.~S9 of the Supplementary Materials,
            it contributes only a sixth of the total predicted heatwave amplitude.
            Now, when we follow the gradient to optimize $S$,
            \begin{equation} \label{eq:pert}
                \frac{\partial}{\partial S}\hat{\mu}_\mathrm{CNN}(S) = M_\mathrm{GA} + \frac{\partial}{\partial S}\hat{\mu}_\mathrm{pert}(S).
            \end{equation}
            Importantly, the first term of the right-hand side does not depend on $S$, while the second one will act differently during the optimization as $S$ evolves and as different initialization seeds $S_0$ are used. It is thus reasonable that once we take the average over all the test set, this second term gets close to 0. However, it will be responsible for the pattern observed in the variance of the optimal input, and thus it draws our attention to the Arctic.

\begin{figure}
              \centering
              \includegraphics[width=0.8\textwidth]{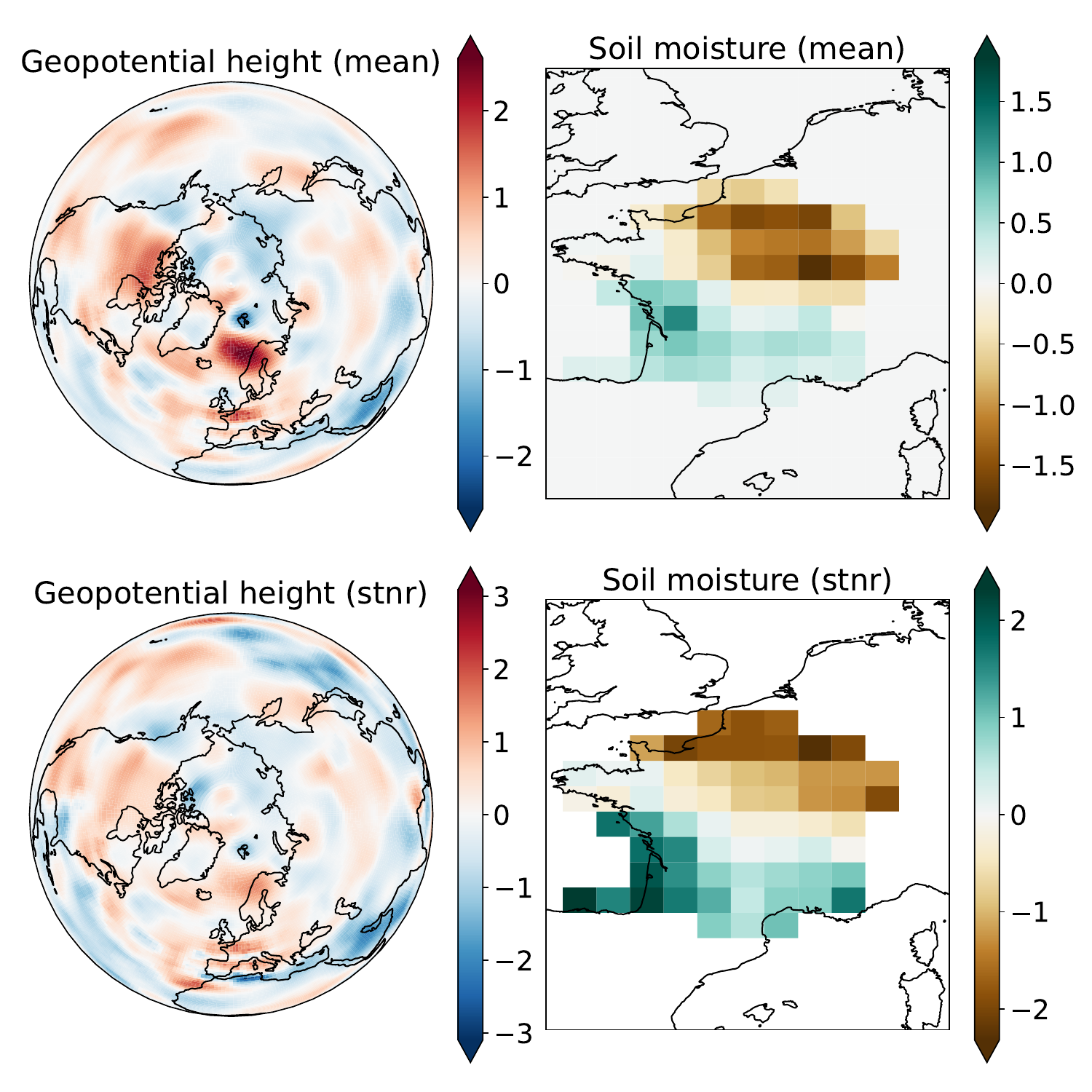}
              \caption{Top: average of the optimal inputs $S$ that maximize the heatwave amplitude $\hat{\mu}_\mathrm{CNN}(S)$ predicted by the CNN while keeping fixed the prediction $\hat{\mu}_\mathrm{GA}(S)$ of the Gaussian approximation, across the different seeds $S_0$ taken from the test dataset. Bottom, signal-to-noise-ratio, computed as the pixelwise ratio between mean (top row) and standard deviation of the optimal inputs (see Fig.~S12 of the Supplementary Materials).}
              \label{fig:oi-orth}
            \end{figure}

            We can investigate the sources of non-linear information more precisely by putting our focus directly on $\hat{\mu}_\mathrm{pert}(X)$, and we can achieve this by adding to the loss $\ell$ another regularization term
            $+\,\lambda_\mathrm{orth} \left( \hat{\mu}_\mathrm{GA}(S) - \hat{\mu}_\mathrm{GA}(S_0) \right)^2$, which strongly nudges the optimization in an orthogonal direction with respect to $M_\mathrm{GA}$.
            Similarly to what we did for $\lambda_2$ and $\lambda_r$, we chose $\lambda_\mathrm{orth} = 10$, which yields the results in Fig.~\ref{fig:oi-orth}, with an average $\hat{\mu}_\mathrm{pert}(S) = 6.1 \pm 0.4 \si{\kelvin}$.

            This time, the average optimal input pattern highlights new interesting features, the first being a North-East to South-West gradient in soil moisture. A possible explanation is that North-Eastern France is generally wetter, so when that region gets dry, the heatwave will be non-linearly more intense (see Eq.~\eqref{eq:pert}).
            The patterns observed in the geopotential height field are harder to interpret, with the most prominent feature being a strong anticyclone over the Norwegian Sea, which further reinforces the findings of \citet{della-martaSummerHeatWaves2007,cassou2005tropical} discussed before.
            Nevertheless, the signal-to-noise-ratio is quite low everywhere, showing that these second order patterns are not very robust.

            Another option to investigate non-linear patterns is to perform K-means clustering on the optimal input maps (both with and without the orthogonality constraint), similarly to what is done in \citet{tomsAssessingDecadalPredictability2021}. Although the distribution of many individual pixels is markedly bimodal (not shown), the maps overall do not appear to be organized in distinct clusters.

            In the end, obtaining global interpretability beyond the linear model proved extremely challenging, so, in the next section, we focus instead on local interpretability.

        \subsubsection{Local interpretability: Expected Gradients} \label{sec:egfi}

            Local interpretability focuses on individual predictions, offering explanations that enhance understanding of feature contributions for each input (e.g., initial conditions leading to heatwaves in our case). Global interpretation techniques often overlook these insights.

            In some cases, then, local interpretability helps to identify the model's strengths, weaknesses, errors and biases, providing feedback for improvement. Methods for achieving local interpretability include feature attribution and counterfactuals. Feature attribution, which includes techniques like saliency maps, assigns values to measure the importance of each input feature. Counterfactuals explain a prediction by examining which features would need to be changed to achieve a desired prediction.

            Several studies have already applied local XAI techniques to explain CNN predictions in geoscience applications \citep[e.g.][]{barnesThisLooksThat2022, tomsAssessingDecadalPredictability2021}, and, in this work, we focus on feature attribution methods, as numerous approaches are available and specifically suited for deep neural networks.
            Gradient-based methods such as Deconvolution \citep{zeiler2014visualizing}, Guided Backpropagation \citep{springenberg2014striving}, and Grad-CAM \citep{selvaraju2017grad} have been developed to perform feature attribution for CNNs. To address the limitations of gradient-based approaches, axiomatic methods such as Layer-wise Relevance Propagation \citep{bach2015pixel}, Taylor Decomposition \citep{montavon2017explaining}, and Deep LIFT \citep{li2021deep} have been introduced. A state-of-the-art axiomatic approach, \emph{Integrated Gradients} \citep{sundararajan2017axiomatic}, adds two new key axioms: sensitivity and implementation invariance.
            We choose the latter approach for these desirable properties and robustness. For a review of various methods and their evaluation in the context of climate science, refer to \citet{bommer2023finding}.

            Consider one input $x$, which in our case is the stack of Z500 and SM fields. Integrated Gradients computes pixel attribution at pixel $i$ as

            \begin{equation} \label{eq:intgrad}
                \text { IntegratedGrads }{ }_i(x) \coloneqq \left(x_i-x_i^{\prime}\right) \times \int_{\alpha=0}^1 \frac{\partial u\left(x^{\prime}+\alpha \times\left(x-x^{\prime}\right); \theta \right)}{\partial x_i} d \alpha
            \end{equation}

            where $u(x;\theta)$ represents one of the outputs of the neural network for which we seek insight; in our case it could be $\hat{\mu}(x;\theta)$, $\hat{\sigma}(x;\theta)$, or directly the heatwave probability.
            $x'$ is a baseline input, such as the mean input on the training set for instance, which is here zero since our data is normalized.
            Importantly, the sum of $\text{IntegratedGrads}_i(x)$ over all pixels $i$ should yield the predicted value $u(x;\theta)$, according to the completeness axiom (see \citet{sundararajan2017axiomatic} for further details). Thus, $\text{IntegratedGrads}_i(x)$, which can be either positive or negative, quantifies the contribution of pixel $i$ to the predicted output $u(x;\theta)$. We used an extension of this method called \emph{Expected Gradients} \citep{erion2021improving}, which reformulates the integral in Eq.~\ref{eq:intgrad} as an expectation and combines that expectation with sampling reference values from a background dataset (taken as the entire training set here). We used the implementation provided by the method \emph{GradientExplainer} in the SHAP package \citep{NIPS2017_8a20a862}.
            Below, we focus on the \SI{500}{\hecto\pascal} geopotential height anomaly input for predicting $u=\hat{\mu}(X)$.
            Results for soil moisture and the prediction of $\hat{\sigma}(X)$ are shown in
            Figs.~S5~and~S6 of the Supplementary Materials.



\begin{figure}[htbp]
                \colorbox{white}{\hspace{-1.60cm} 
                \includegraphics[scale=0.45]{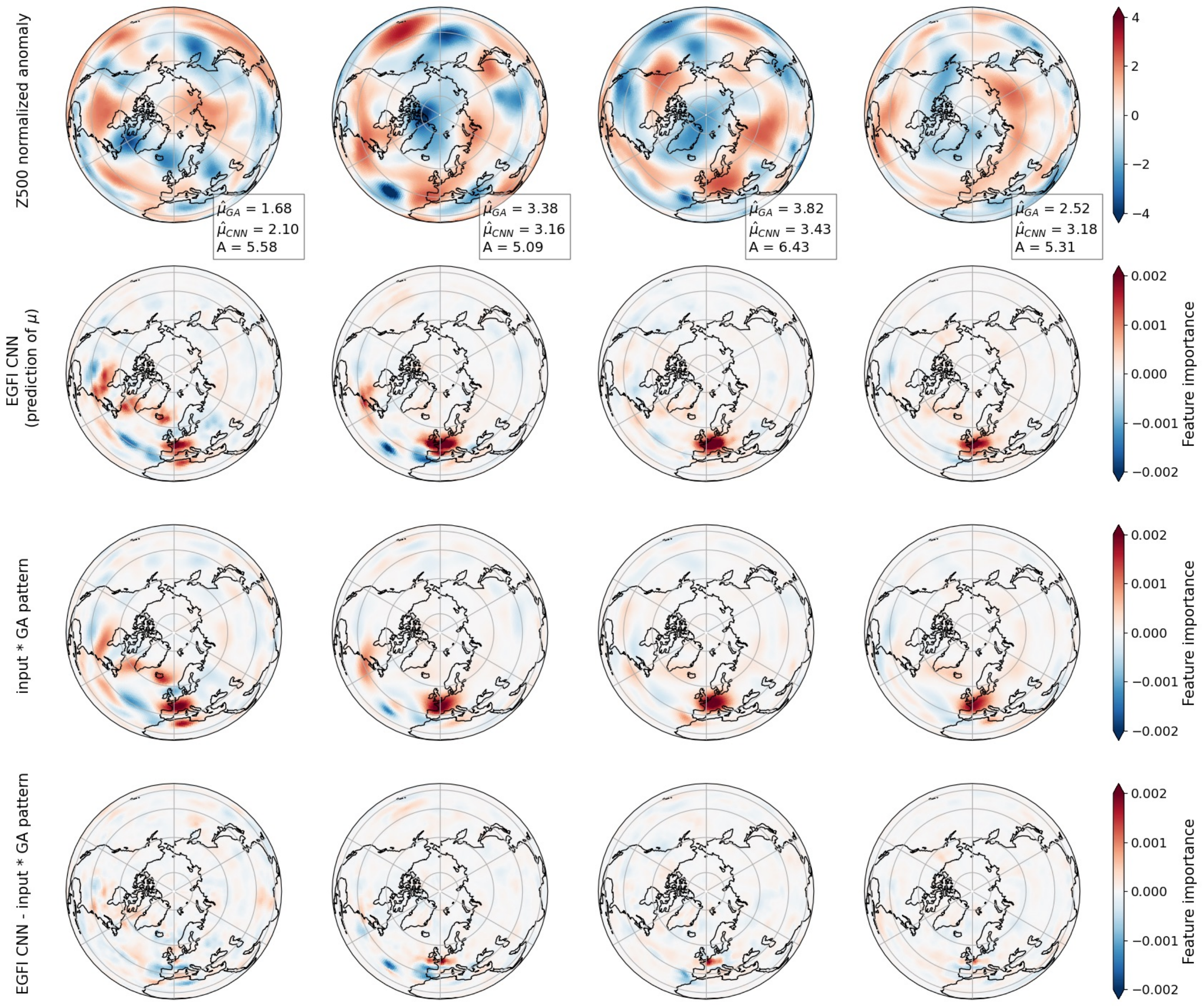} 
                \hspace{1cm}}
                \caption{Top row: Several normalized Z500 (no units) initial conditions $X$ associated with $A$ above the $99^{th}$ percentile (heatwaves). Second row: Expected Gradients feature importance (EGFI) of the CNN predictions on these inputs. Third row: pointwise multiplication between inputs and GA projection pattern. Since $\hat{\mu}_{GA}(X)$ is linear in $X$, this is equivalent to EGFI for the GA prediction. Fourth row: EGFI CNN minus EGFI GA for each input.}
                \label{fig:shap_local_FI}
            \end{figure}

            To illustrate the feature importance derived from the CNN predictions on relevant cases, we randomly selected four examples from the 100 most extreme heatwaves in the test set.
            The corresponding Z500 initial conditions are shown in Fig.~\ref{fig:shap_local_FI} (first row), together with the resulting Expected Gradients feature importance (EGFI) maps for the CNN prediction (second row).
            In practice, EGFI maps are smoothed by a Gaussian filter for visual purposes (see
            Fig.~S3 of the Supplementary Materials).

            For such extreme events, the heatwave amplitude $A$ was already above the 5\% threshold $a_5 = \SI{3.11}{\kelvin}$ for several days, and the main dynamics is that of an anticyclone over Western Europe progressively getting stronger \citep[see for instance][]{zschenderleinProcessesDeterminingHeat2019}.
            It is thus natural that most examples (among the ones shown all but the first) exhibit a prominent positive feature over France.
            The feature importance maps also reveal the role of some regions in the North Atlantic and over North America for the CNN prediction.
            Positive geopotential anomalies in the storm tracks region are consistently associated with positive contribution to predicted temperatures (in particular in examples 1 and 2), as well as negative anomalies over the Labrador sea or Icelandic lows (see in particular example 1).
            Another region of importance is the Eastern Atlantic Ocean: anticyclonic anomalies off the coast of Portugal contribute negatively to predicted temperatures (examples 1, 2 and 4).
            This behavior could be related to the recently highlighted role of deep depressions in this region for extreme temperature over Western Europe~\citep{Faranda2023,DAndrea2024}.

            As we are interested in understanding what the CNN has learned beyond the GA, we compared the EGFI maps of the CNN with the EGFI maps of the GA.
            As the GA model predicts $\hat{\mu}(X)$ linearly, its EGFI maps are exactly equal to the input $x$ multiplied by the GA projection pattern $M_\mathrm{GA}$; we show these maps in the third row of Fig.~\ref{fig:shap_local_FI}.
            The comparison reveals that the patterns of feature importance are strikingly similar, with the regions of importance largely overlapping.

            This suggests that the features described above correspond to some regions of the large-scale patterns identified in the projection maps in section~\ref{sec:ga-projmap}; although no single individual Z500 anomaly snapshot exhibits such a large-scale coherent structure, they all project to some extent on this pattern.
            Looking at the difference between the EGFI maps of the GA and those of the CNN (Fig.~\ref{fig:shap_local_FI}, fourth row), we observe that the additional contribution leading to the prediction of the CNN behaves differently in different geographical locations. For instance, over France, and to a lesser extent in the storm tracks region, the CNN systematically reinforces the weight already identified by the GA. In other regions, like the Icelandic low, it systematically tones down the signal seen in the GA. And finally, other locations, like off the coast of Portugal, show a different sign depending on the particular input.
            In most cases, this correction that the CNN makes on the GA baseline yields more localized CNN patterns. This suggests that, due to its nonlinearity and convolutional nature, the CNN model might be capturing more complex interactions over multiple scales, that the GA model, due to its linear nature, cannot represent.

            While extracting detailed information from these plots is challenging due to the qualitative nature of this local input-dependent approach, we have quantified the correlation between the EGFI maps and the GA maps more systematically for all inputs in the test. We obtained an average correlation of $0.84 \pm 0.08$ (full histogram in
            Fig.~S4 of the Supplementary Materials),
            which confirms the fact that the EGFI maps of GA and CNN are indeed consistently similar, not only for the very extreme events shown in Fig~\ref{fig:shap_local_FI}.

            Additionally, a global feature importance map can be generated by averaging the absolute values of the local maps across the entire dataset.
            However, like the \emph{optimal input} method presented in the previous section, this approach would be limited, as it would only highlight the areas of the input deemed important by the CNN model, not how this information is used to make predictions \citep{rudinStopExplainingBlack2019}.

            A particularly relevant question is which scales are significant and how different scales interact.
            An ad-hoc attempt to address this issue for any black-box model (including CNNs) is discussed in \citet{kasmi2023assessment}.
            In the next section, we partially address this question with the ScatNet model.

    \subsection{ScatNet} \label{sec:interp-scatnet}




        At the end of the scattering transform, we obtain a feature map with the shape $\left(\frac{N_{lat}}{2^J}, \frac{N_{lon}}{2^J}, 1 + J \times L \right)$, where $N_{lat}$ and $N_{lon}$ are the number of latitude and longitude points in the input image, $J$ is the number of scales, and $L$ is the number of orientations. The first feature map corresponds to the zeroth-order scattering transform, which is just the low-pass-filtered input.
        The subsequent $J \times L$ feature maps correspond to the first-order scattering transform. These maps are generated by convolving the input field with wavelet filters at each scale $j \in \{0,1,2\}$ and orientation $\theta \in \{0, \ldots, 7\}$ (since $J=3$ and $L=8$), followed by applying the modulus operator and pooling \citep[]{brunaInvariantScatteringConvolution2013}.
        Similarly to the GA and IINN models, we can project the learned weights onto spatial maps for each channel of the feature maps.
        For the weights to be proper (global) \emph{feature importance} maps, we normalize them to account for the different ranges of values of each feature map after the scattering transform.
        Namely, the feature importance \( FI_i\) for each feature \( i \) with associated weight $\beta_i$ in the final linear layer of the network is given by
        \begin{equation}
            FI_i = \E(\left| X_i - \E(X_i) \right|) \cdot \beta_i.
        \end{equation}
        The expectation is taken as the mean over the test dataset.
        This definition corresponds to the mean absolute value of EGFI for the ScatNet model (see section~5.\ref{sec:egfi}) for feature \( i \), but retaining the information of the sign of the weight.

\begin{figure}[htbp]
            \centering
            \includegraphics[width=0.8 \textwidth]{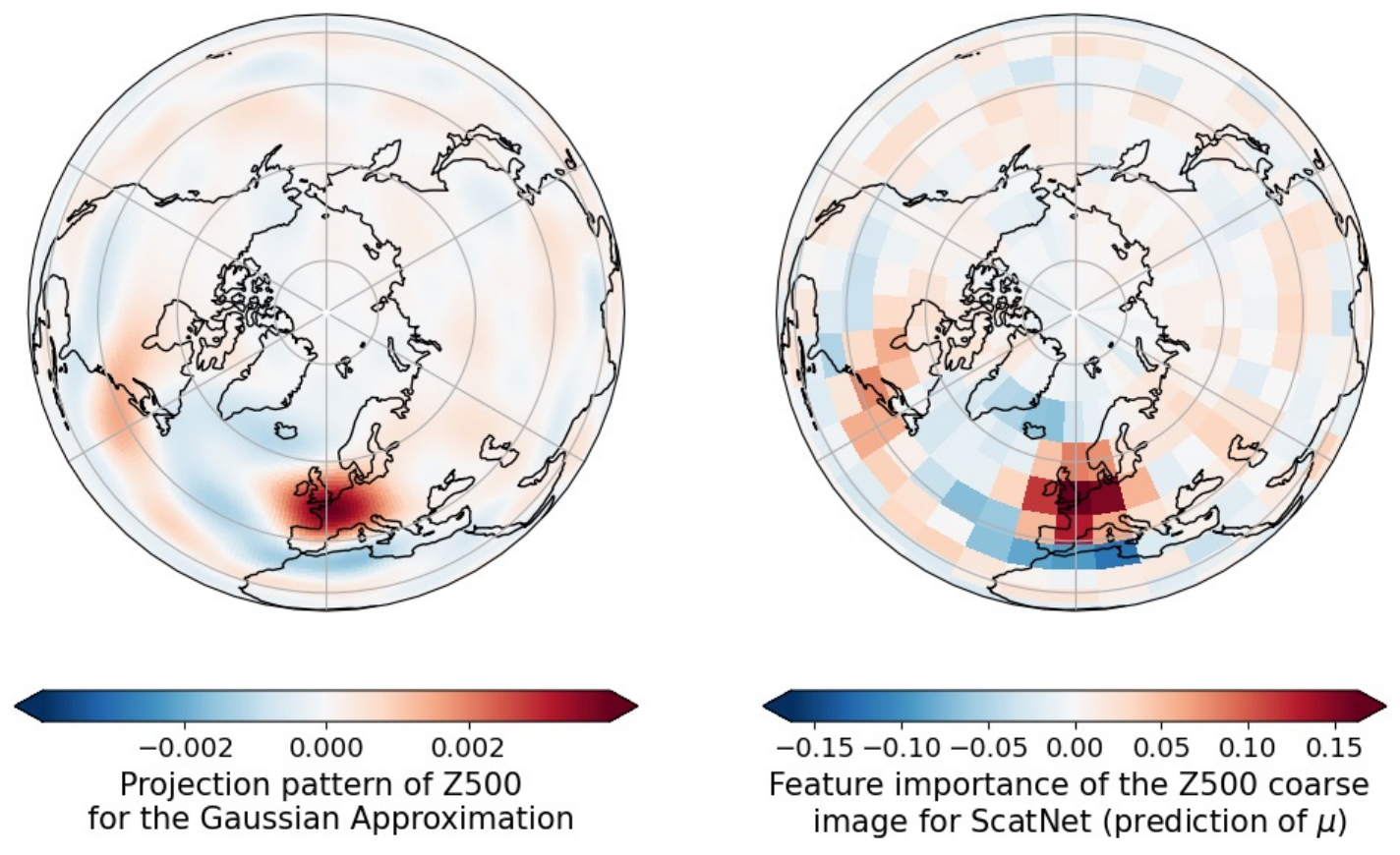}
            \caption{Projection patterns of the GA (left) and the feature importance of coarse Z500 field for the prediction of $\hat{\mu}(X)$ with ScatNet (right). We show, among the 5 models, the one with the highest skill.}
            \label{fig:scatnet-0order-pattern}
        \end{figure}

        Here we focus on interpreting the scattering features of the geopotential height anomaly at \SI{500}{\hecto\pascal} (Z500) for the prediction of $\hat{\mu}(X)$. We first examine the feature importance maps corresponding to the zeroth order features (or \emph{coarse field}). These are obtained by simply applying a Gaussian low-pass filter to the input Z500 field, allowing us to directly compare the resulting projection pattern with the one computed by the GA model, as shown in Fig.~\ref{fig:scatnet-0order-pattern}.
        The two patterns are nearly identical, up to the coarsening. This is not surprising, but rather reassuring. It can be interpreted as the fact that coarse-graining by a factor of $2^J$ acts as a regularization, similar to how we smoothed the GA projection pattern.

        A natural question is whether we can achieve a predictive performance similar to the GA model with a lower resolution Z500 field. It amounts to know which is the effective scale for a linear prediction.
        To answer this question, we define ScatNet$_{\mathrm{coarse}}$, a variant of ScatNet where only the zeroth order features (the coarse Z500 field) are computed, rather than proceeding to the first order in the scattering transform.
        In Tab.~\ref{tab:skill_vs_LR_coarse}, we compare its predictive performance with the GA and CNN models.
        \begin{table}[htbp]
            \centering
            \begin{tabular}{c|c|c|c|c|}
                \multicolumn{2}{c}{} & \multicolumn{3}{c}{Metric} \\
        		\cline{3-5}
                \multicolumn{2}{c|}{} & CRPSS & NLLS & BCES \\
                \cline{2-5}
                \multirow{4}{*}{\rotatebox[origin=c]{90}{Model}}
                & GA &$0.2864 \pm 0.0009$ & $0.2169 \pm 0.0009$ & $0.293 \pm 0.001$ \\
                \cline{2-5}
                & ScatNet$_{\mathrm{coarse}}$ &$0.2862 \pm 0.0005$ & $0.203 \pm 0.001$ & $0.291 \pm 0.001$ \\
                \cline{2-5}
                & CNN &$0.310 \pm 0.003$ & $0.245 \pm 0.007$ & $0.311 \pm 0.008$ \\
                \cline{2-5}
            \end{tabular}
            \caption{Test skills (the higher, the better) of the different models. We compare ScatNet$_{\mathrm{coarse}}$ to our most simple approach (GA) and our more complex one (CNN).}
            \label{tab:skill_vs_LR_coarse}
        \end{table}
        We observe that ScatNet$_{\mathrm{coarse}}$ achieves a performance barely lower than the GA model that uses the $Z500$ field at a higher resolution. This suggests that the performance gains from the ScatNet and CNN models are due to processing finer-scale information.

        To further validate this statement, Tab.~\ref{tab:importance-vs-scales} compares the relative global feature importance for different scales of the Z500 coarse field, and the pixels of SM.
        These values are derived by calculating the feature-wise mean of the absolute feature importance values across the entire test dataset and then, independently for each scale, summing over geographical locations and different orientations.
        \begin{table}[htbp]
            \centering
            \begin{tabular}{c|c|c|c|c|c|c|}
                \cline{3-7}
                \multicolumn{2}{c|}{} & scale $j=0$ & scale $j=1$ & scale $j=2$ & coarse field & soil moisture \\
                \cline{2-7}
                \multirow{2}{*}{} & Relative FI (in \%) & $5.2 \pm 0.3$ & $10.4 \pm 0.5$ & $20.0 \pm 1.0$ & $51.0 \pm 2.4$ & $13.5 \pm 1.0$ \\
                \cline{2-7}
            \end{tabular}
            \caption{Relative feature importance of various scales, expressed as percentages. The first three columns represent the relative mean absolute feature importance on first-order feature maps, summed across all orientations for scales $j=0, 1$ and $2$. The fourth column shows the mean feature importance on the zeroth-order Z500 feature map. The final column presents the mean feature importance summed across all pixels of soil moisture in France. We show the average values across the 5 models.}
            \label{tab:importance-vs-scales}
        \end{table}
        We observe that nearly 65\% of the feature importance is attributed to the coarse Z500 field and soil moisture, i.e. the information accessible to the GA model. The coarse Z500 field cannot distinguish structures occurring below $2^3$ pixels (approximately \SI{800}{\kilo\meter} at 45 degrees North), which corresponds roughly to the synoptic scale. It is then not surprising that most of the predictive power arises from above or equal to this scale. We also note that the soil moisture projection pattern is quite noisy (not shown) due to the absence of regularization, yet it still exhibits relatively high feature importance.
        The remaining 35\% of feature importance is derived from the first-order scattering features, with larger scales contributing more. During the scattering transform process, when we convolve our Z500 field with a wavelet filter located at scale $j$, we are essentially applying a band-pass filter concentrated around the wavelength of $2^j$ pixels. However, this band-pass filter is non-local in Fourier space, meaning that it affects a range of wavelengths rather than a single one, and thus the different scales are not completely independent.
        Nevertheless, Tab.~\ref{tab:importance-vs-scales} indicates that additional information at the sub-synoptic scale (\SI{400}{\kilo\meter} and below) has an important role for prediction. \\

\begin{figure}[htbp]
            \colorbox{white}{\hspace{-1cm} 
            \includegraphics[width=1.1\textwidth]{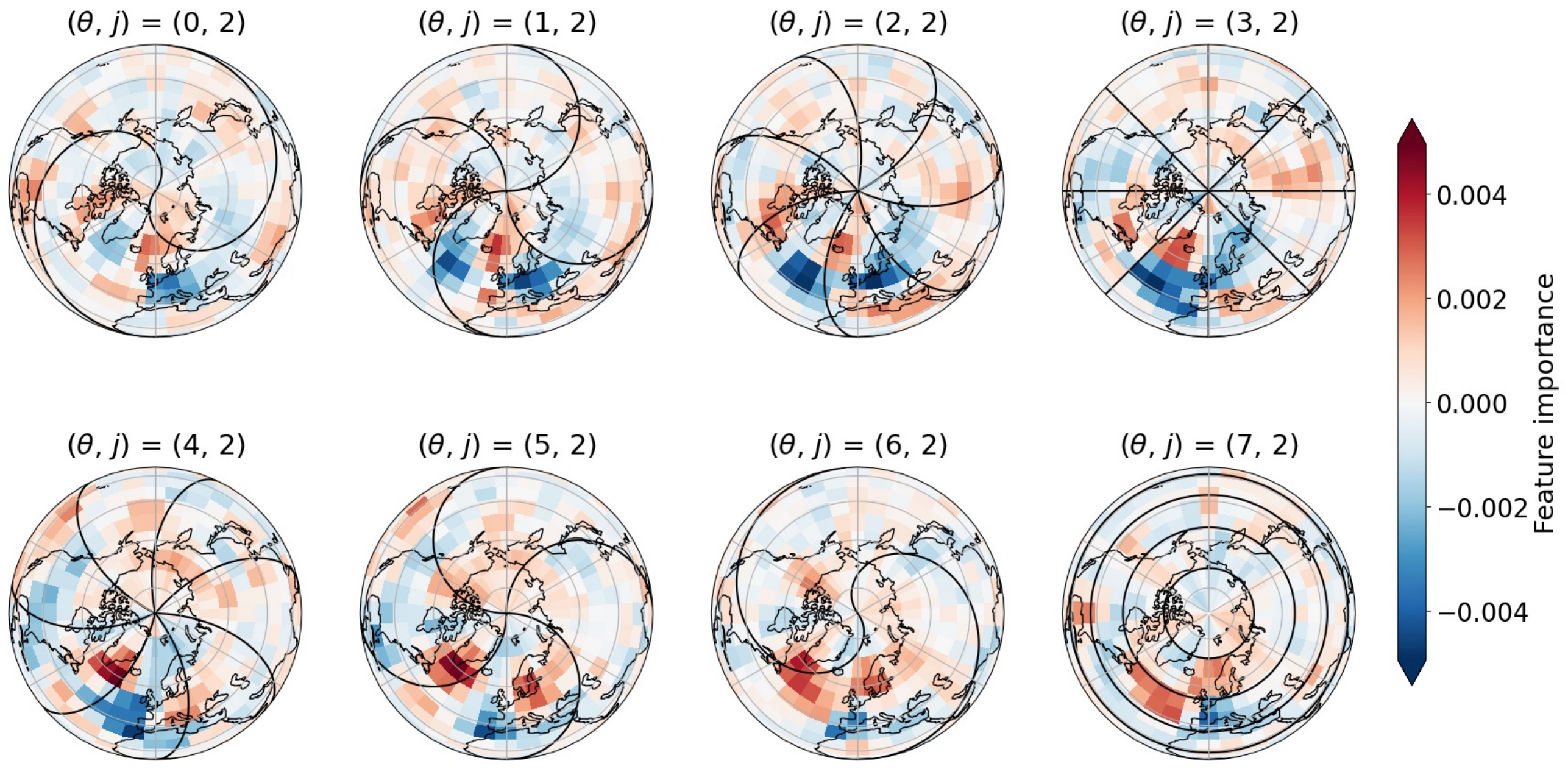} 
            \hspace{0.8cm}}
            \caption{Feature importance of first order features for the prediction of $\hat{\mu}(X)$ with ScatNet at scale $j=2$ for each orientation. Maps at finer scale are very similar, but with less feature importance (see Fig.~S13 of the Supplementary Materials). The black filaments represent the orientation of the wavelet in spatial space, with the approximate wave vector being orthogonal to the filaments. We show the average maps across the 5 models.}
            \label{fig:1st-order-patterns}
        \end{figure}

        Finally, we present in Fig.~\ref{fig:1st-order-patterns} the feature importance patterns derived from all the first-order scattering features at scale $j=2$. For scale $j=0$ and $j=1$, the feature importance maps are very similar, but with less feature importance
        (see Fig.~S13 of the Supplementary Materials).
        To interpret these maps, let us consider for example the map corresponding to $\theta=3$ (top right). In the tiling of the Fourier space, $\theta=3$ corresponds to a wave vector oriented horizontally, and thus to a zonal oscillation. The blueish structure in the Atlantic Ocean indicates that zonal oscillations at a scale of $2^2$ pixels in this location contribute to a decrease in the predicted value of $\hat{\mu}(x)$, and consequently, to a lower probability of a heatwave.
        Another relevant case is that of the map at $\theta=7$ (bottom right), which corresponds to meridional oscillations at a scale of $2^2$ pixels. The reddish structure, located, as in the previous example, in the middle of the Atlantic Ocean, indicates this time that these oscillations increase the likelihood of a heatwave.
        It should be noted that areas exhibiting the same hue in this figure do not represent structures corresponding to the size of the color segments. Instead, they denote regions in which structures approximately \SI{400}{\kilo\meter} in scale produce equivalent effects, irrespective of their exact location within the region, with darker shades signifying a more pronounced impact.
        We observe that the patterns appear to evolve continuously from one orientation to another. This can be partially explained by the overlap of the Fourier supports of wavelet filters from one orientation to the next \citep{brunaInvariantScatteringConvolution2013}.
        In terms of physical interpretation, we first note that most of the feature importance is concentrated around the target region (France) and upstream regions in the Atlantic Ocean.
        Many of these regions coincide with those identified before, for instance Scandinavia, Iceland, the storm tracks and the Labrador sea, as well as the Eastern Atlantic close to the coast of Portugal.
        This confirms our previous findings that additional skill beyond linear regression essentially comes from refinement of patterns already identified in the optimal projection.
        One of the main benefits of the scattering transforms is that it allows us to identify the scales and orientations associated with those refinements, and not just their location.
        In particular, we see that in some regions the feature importance always has the same sign for all orientations: negative over France, positive over Iceland.
        On the other hand, most of the other regions are sensitive to the orientation of the disturbances.
        For instance, North-South oscillations in the East Atlantic promote heatwaves while East-West ones in the same area inhibit them.


\section{Discussion}

By testing increasingly complex architectures, we were able to quantify the extent to which higher complexity leads to a better performance. As pointed out in \citet{rudinStopExplainingBlack2019}, the benefits of complexity are sometimes overrated, and indeed, when data is scarce, the best performance was achieved by the simple linear regression of the Gaussian approximation.
This simple model remained competitive with the more complex ones, even with much larger amounts of data. Moreover, when the Scattering Network uses only the coarse-grained input, it becomes itself a simple linear regression, and it has the same skill as the Gaussian approximation while using 64 times fewer scalar features. This highlights the potential for even further simplification of the regression task without loss of performance from the GA baseline.

For the task of forecasting heatwaves, the slightly increased complexity of the Intrinsically interpretable Neural Network did not give any benefits, neither from the point of view of performance nor of additional insight into the physical processes.
However, the design of the architecture is very interesting, reimagining a simple fully connected neural network with a bottleneck in terms of optimal linear projection of the data. It is then easy to relax the bottleneck from a single optimal index to a set of $m$ optimal indices. This could be particularly useful when studying problems where the effective dynamics can be well expressed non-linearly in a low-dimensional space. A possible example could be the study of the Madden-Julian Oscillation \citep{delaunayInterpretableDeepLearning2022}. Compared to standard principal component analysis, the IINN would give linear projections that are tailored to the regression task, rather than the ones that best explain the variance of the input data.
We did try to train IINN architectures with more than one projection pattern on the heatwave prediction task, but the results were not better than the ones presented in this work.

Going beyond the skill of the linear model requires several centuries of data to properly train more flexible architectures, like Convolutional Neural Networks.
With post-hoc explainability methods, we found that the CNN essentially refines and modulates the projection pattern of the linear model.
However, we advocate for caution in using these methods. If we had just looked at the CNN and its post-hoc explanations, we would have been happy to see them highlighting dry soil, the anticyclone over France and activity over the Atlantic.
Only by having a hierarchy of models we realize that these easily recognizable features were already present in the linear model, and thus the \emph{extra} skill that we gained lies in second order details.
Pinpointing such details is not easy, and we could not draw robust conclusions.
Nevertheless, the explainability maps do provide \emph{hints} of where extra sources of predictability \emph{might be}, which can be useful for future investigations with more rigorous methods.


The architecture of the ScatNet proposed in this work allows to overcome the limitations of post-hoc tools, providing global interpretability like GA and IINN. Importantly, the ScatNet has the same skill as the CNN, so we can say that there is nothing significant in the predictions of the CNN that the ScatNet misses.
Our analysis shows that the linear model contributes to 65\% of the prediction of the ScatNet, with the remaining information being encoded in oscillations in the \SI{500}{\hecto\pascal} geopotential height anomaly field at sub-synoptic scales. Even more interestingly, we are able to visualize the individual contribution of oscillations in different directions and geographical locations.
The precise physical interpretation of these oscillations goes beyond the scope of this paper, but we feel this is a very promising direction for the discovery of new physics.

As a technical note, in this work we apply the Scattering Transform to the geopotential height field, treating it as a 2D image. However, a spherical adaptation \citep{mcewen2021scattering} would better account for Earth’s geometry. ScatNet could then serve as an interpretable preprocessing step, providing a sparse representation of geophysical fields—either as an encoder’s initial layer or as a regularization term to enforce physically consistent forecasts. Meanwhile, our Gaussian approximation of the conditional distribution of heatwave amplitude, $\mathbb{P}(A|X)$, may limit performance in nonlinear cases where distributions are skewed or multimodal. Other approaches like quantile regression \citep{zhang2018improved} or extreme value theory \citep{cisneros2024deep} could improve flexibility but at the cost of increased model complexity and reduced interpretability, which was our primary focus. \\


From a more physical standpoint, in this paper we call heatwaves \emph{any} points for which $A(t) \geq a_5$, which naturally means our heatwave samples are correlated, and often belong to a longer-lasting or more instense heatwave. This means we are more sensitive to stationary features, characteristic of the \emph{continuation} of a heatwave rather than its \emph{onset}. Similarly, as for any statistical model, finding a prediction that works well for all the stages of a heatwave tends to average out the dynamics. Ideally one could train different models on the different stages, but that comes with the issue of precise definitions of said stages and reduced dataset size.
We tried a-posteriori to look at the EGFI maps of the CNN specifically at heatwave onsets. If onset dynamics are very different from continuation, the CNN, being highly non-linear, should be able to capture them both. However, there was no obvious difference with respect to EGFI maps captured at the center of long lasting heatwaves.

This points out that explainable AI methods are only the first step in understanding the problem at hand.
Once interesting features are identified, we should complement the analysis by computing composite statistics, conditioned on the occurrence of those features rather than the target event itself.
Alternatively, we could perform denial experiments, where those features are either the only information a network can see or are instead masked out.
This would allow to precisely and reliably quantify their effective importance for prediction.
Another very interesting possibility would be to run idealized experiments with climate models, where, for instance, we explicitly force the sub-synoptic oscillations found by the ScatNet.
This would require first some analysis work to identify the conditions which favors those oscillations.
These experiments are beyond the scope of this paper, which was to investigate the tradeoff between performance and \emph{interpretability}.
However, we think the results presented in this work will be useful to improve, in future studies, their \emph{interpretation}.


\section{Conclusions}



In this work, we tested three increasingly complex interpretable models and a black-box one on the task of forecasting extreme heatwaves.
Our findings can be summarized in three main conclusions: first, when data is scarce, the simplest model (GA) has the highest skill.
Second, even with enough data the black-box model (CNN) does not outperform the best interpretable model (ScatNet). Moreover, its explanations are small modulations on what was known from the GA model, and thus are insufficient to pinpoint the source of its extra skill.
Third, the ScatNet model identifies the scale and orientation of the features relevant for prediction, allowing us to easily find that the gain in performance from GA comes from oscillations in the geopotential height field, mainly over the North Atlantic and with a wavelength around \SI{400}{\kilo\meter}.

However, these conclusions hold for this specific problem, where nonlinearity is moderate; for more nonlinear tasks, black-box models like CNNs could be more advantageous.

With this work, we highlighted once again the power of interpretable models, and we identified Scattering Networks (ScatNets) as a very promising tool to be applied in climate science.
We showed that statistical models learned known drivers of extreme heatwaves, making quantitative statements about them. Also, the different interpretability tools used in this work point at where new sources of predictability may be found, with the most prominent being the sub-synoptic oscillations in the geopotential height field mentioned above.
It is then a natural perspective to investigate them from a more dynamical point of view and hopefully gain a mechanistic understanding.
Finally, the methods used in this study could be extended to observational datasets or model output with a dynamical ocean, which would allow looking for heatwave drivers beyond land and atmospheric variables.

\clearpage
\acknowledgments
Author Lovo received funding from the European Union’s Horizon 2020 research and innovation program under the Marie Skłodowska-Curie grant agreement 956170 (Critical Earth).
Author Lancelin is funded by RTE France, the French transmission system operator, and benefited from CIFRE funding from the ANRT. This work is also supported by the Institut des Mathématiques pour la Planète Terre (IMPT). The authors gratefully acknowledge the support of this project.
The authors also warmly thank Stéphane Mallat and Nathanaël Cuvelle-Magar for fruitful discussions about the Scattering Transform, as well as to Laurent Dubus (RTE) for his insightful and constructive feedback on the manuscript.

The authors thank the reviewers for their comments that helped improve this paper.

%
%
\datastatement
Code is available on Zenodo (\url{https://doi.org/10.5281/zenodo.14922107}), together with all the data needed to reproduce the results in this paper. Due to its size, it was not possible to publish the 1000-year-long control run of CESM. However, the details and code of how this control run was obtained are explained in \citet{ragoneRareEventAlgorithm2021} and available on Zenodo (\url{https://doi.org/10.5281/zenodo.4763283}).

%






%



\bibliographystyle{ametsocV6}
\bibliography{actual}

\end{document}